\documentclass{article}
\usepackage{spconf,amsmath,graphicx}

\usepackage{svg}
\usepackage{amssymb}
\usepackage{multirow}
\usepackage{booktabs}
\usepackage[ruled,vlined]{algorithm2e}
\newcommand{\STAB}[1]{\begin{tabular}{@{}c@{}}#1\end{tabular}}
\DeclareMathOperator*{\argmax}{\arg\!\max}

\title{Ada-SISE: Adaptive Semantic Input Sampling for Efficient Explanation of Convolutional Neural Networks}
\name{%
\begin{tabular}{@{}c@{}}
Mahesh Sudhakar$^{\star}$, Sam Sattarzadeh$^{\star}$, Konstantinos N. Plataniotis$^{\star}$, 
\\
Jongseong Jang$^{\dagger}$, Yeonjeong Jeong$^{\dagger}$, Hyunwoo Kim$^{\dagger}$
\end{tabular}}  
  
\address{$^{\star}$Department of Electrical \& Computer Engineering, University of Toronto\\ $^{\dagger}$Fundamental Research Lab, LG AI Research}

\begin{document}
%
\maketitle

\begin{abstract}


Explainable AI (XAI) is an active research area to interpret a neural network's decision by ensuring transparency and trust in the task-specified learned models. Recently, perturbation-based model analysis has shown better interpretation, but backpropagation techniques are still prevailing because of their computational efficiency. In this work, we combine both approaches as a hybrid visual explanation algorithm and propose an efficient interpretation method for convolutional neural networks. Our method adaptively selects the most critical features that mainly contribute towards a prediction to probe the model by finding the activated features. Experimental results show that the proposed method can reduce the execution time up to 30\% while enhancing competitive interpretability without compromising the quality of explanation generated.

\end{abstract}


\section{INTRODUCTION}
\label{sec:intro}

Over the recent past years, access to a lot of digital data, the advances in computing facilities, and the facile access to many readily available pre-trained models have fueled the growth in deep learning. Although such models produce high accuracy in object recognition, the interpretability \cite{fan2020interpretability} of their decisions is also essential to convince the stakeholders or locate any potential bias in the underlying data. With AI currently being employed in various fields such as in healthcare, consumer retails, and banking, it is high time to develop ``Responsible AI" \cite{arrieta2020explainable} for society. To ensure the uniformity of the training data's distribution, lately, there is an increase in modern open-source toolkits \cite{adebayo2016fairml, bellamy2019ai} that acts as a common framework to evaluate a model's fairness.



Explainable AI (XAI) has recently been offering many algorithms to interpret a model's behavior. Based on their usage at the training process's timeline, XAI approaches can be broadly classified into \textit{ad-hoc} and \textit{post-hoc} methods. In terms of their explanation ability to interpret a single instance or the whole decision process, XAI can be classified into \textit{local} and \textit{global}. They can also be categorized into \textit{model-agnostic} and \textit{model-specific} methods, based on the requirement to specify the model's architecture.

In this work, we study such a recent \textit{post-hoc}, \textit{local}, and \textit{model-specific} XAI algorithm - Semantic Input Sampling for Explanation (SISE) \cite{sattarzadeh2020explaining} developed for image classification tasks. Building on this method, we propose a way to improve its run-time while retaining its overall performance without compromising the visual explanation's quality. Our approach introduces a novel way to adaptively select the most important feature information to be considered for the subsequent steps of the algorithm's operation. This modification acts as a smart filtering procedure that mutates the existing method into an automated, unified solution by eliminating the need for an end-user to tune the hyper-parameters. To demonstrate this claim, we evaluate our approach with the original algorithm's performance in terms of the visual explanation quality, overall benchmark analysis, and execution time. 

\section{BACKGROUND}
\subsection{Existing methods}

The prior works on \textit{post-hoc} visual XAI can be divided into three main groups: `backpropagation-based', `perturbation-based', and `CAM-based' methods. The backpropagation-based methods mainly operate by backpropagating the signals from the output neuron of the model to the inputs \cite{IntegGrad, simonyan2013deep} or the hidden nodes of the model \cite{srinivas2019full}, in order to calculate gradient \cite{simonyan2013deep} of relevance \cite{bach2015pixel} terms. Perturbation-based approaches rely on feed-forwarding the model with perturbed copies of the input image. They interpret the model's behavior using techniques such as probing the model with random masks \cite{petsiuk2018rise} or optimizing a perturbation mask for each input \cite{fong2019understanding}. Moreover, CAM-based methods are built based on the Class Activation Mapping (CAM) method \cite{CAM} and are used specifically for CNNs by taking advantage of the phenomenon of this type of networks in weak object localization, as stated in \cite{zhou2014object}.  Most of these methods are developed by backpropagation techniques \cite{GradCAM, GradCAMPP} or perturbation techniques \cite{ScoreCAM}.

\subsection{Semantic Input Sampling for Explanation}

SISE is a recent explanation method that spans among all three mentioned visual XAI methods, although it is generally classified as a perturbation-based algorithm. SISE employs the feature information underlying the model's various depths to generate a saliency-based high-resolution visual explanation map. For a given trained classification model $\delta: I \rightarrow \mathbb{R}^{C}$
with $N$ convolutional blocks that outputs a confidence score over $C$ classes for each input image $I$, SISE generates a 2-dimensional explanation map $Y_{I,\delta(\lambda)}$ for $\lambda$ in the domain of feature maps $\Lambda$, 
through its four-phased architecture. 






  


In the first phase (\textit{Feature map Extraction}), pooling layers $p_{l}$ of the model for $l \in \{1,..,N\}$ are targeted, and their corresponding feature maps $F_{k}^{[p]}$ for $k \in \{1,..,M^{p}\}$ are collected. As this operation is independent of the classifier part, there would be a lot of irrelevant feature information about the background or other object classes (if present), in addition to the class of interest. The excess information is filtered out in the second phase (\textit{Feature map Selection}) based on their backpropagation scores. Here, the feature maps with positive gradients towards a particular class 
are selected and post-processed to be converted into attribution masks $A_{k}^{[p]}$, via bilinear interpolation followed by normalization in the range $[0,1]$. 

The generated attribution masks are then scored by weighing based on their classification scores in the third phase (\textit{Attribution mask Scoring}) and later combined to form a layer visualization map $V_{I, \delta(\lambda)}^{[p]}$. These preceding steps are repeated for all pooling (down-sampling) layers $p_{l}$ of the network and then passed to the final phase (\textit{Fusion}) of the algorithm, where they are fused in a cascading manner under a series of operations including addition, multiplication, and adaptive binarization, to reach the final explanation map.  

\section{PROBLEM STATEMENT}

The gradient-based feature map selection policy in SISE is aimed to distinguish the feature maps containing essential attributions for the model's prediction (`positive-gradient') against the ones representing outliers or background information. That was achieved using a threshold parameter $\mu$ that was set to 0 by default to
discard the ones with `negative-gradient' scores.

\begin{figure}[ht]
     \centering
     \includegraphics[width=0.8\linewidth]{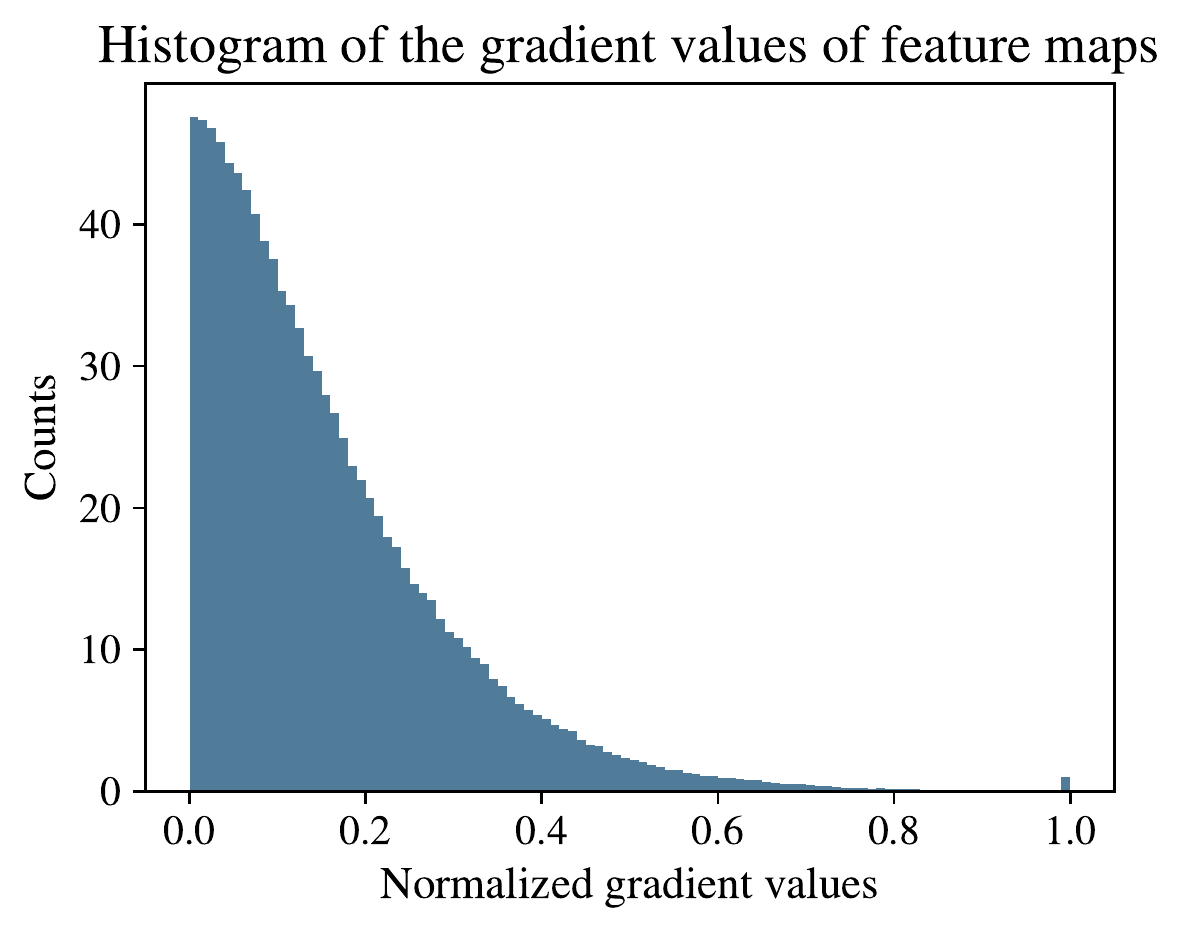}
     \caption{Histogram of the gradient values recorded from the feature maps in the last convolutional layer of a ResNet-50.}
     \label{fig:hog}
 \end{figure}


However, most of the elected activation maps with positive gradients are relatively ineffective in the prediction procedure, thereby increasing SISE's computational overhead unnecessarily. We identify that the average gradient distribution of the positive-gradient feature maps follows a pattern, as in Fig. \ref{fig:hog}, where several trivial features are represented with low gradient values. Thus, only a fraction of the most critical feature maps is passed to the third phase of SISE. Hence, we focus on developing an adaptive selection policy for the parameter $\mu$ of SISE to estimate the least number of required features to generate an explanation map without any notable compromise (and even in some cases, a slight enhancement) in terms of visual quality.

\section{ADAPTIVE MASK SELECTION}
\begin{algorithm}[ht]
\SetAlgoLined
\SetKwInOut{Input}{Input} 
\SetKwInOut{Output}{Output}
\SetKwData{Parameter}{Parameter}
\Input{An input image $I$ and a trained model $\delta$.}
$\eta \leftarrow$ \text{post-processing function.}\\
$\zeta \leftarrow$ \text{heatmap fusion function.}

\For{$n \leftarrow 1,...,N$}{
  Select the pooling layer $p$ and collect feature maps $F_{k}^{[p]} \forall k \in \{ 1,..,M^{p} \}$\;
  Let the domain of the feature maps be $\Lambda^{[p]}$\;
  $\sigma_{k}^{[p]}$ = $\displaystyle\sum_{\lambda^{[p]} \in c} \frac{\partial \delta(I)}{\partial F_{k}^{[p]} (\lambda^{[p]})}$ $\And$  $\rho^{[p]}$ = $\max(\sigma_{k}^{[p]})$ \;
  $A_k^{[p]} \leftarrow []$ ; $\Upsilon^{[p]}\leftarrow\{\upsilon_k^{[p]} > 0 | k\in\{1,...,M^{[p]}\}$\;
 $\mu^{[p]}\leftarrow \Upsilon^{[p]}(\argmax_{j\in \{1,...,|\Upsilon^{[p]}|\}}(\tau^{[p]}(j)))$\;
\ForEach{$k \leftarrow \{1,...,m^{p}$\}}{ \eIf{$\frac{\sigma_{k}^{[p]}}{\rho^{[p]}}>\mu^{[p]}$}{
  $A_{k}^{[p]}\leftarrow A_{k}^{[p]} \cup \eta (F_{k}^{[p]}) $\;
   }{
  $A_{k}^{[p]}\leftarrow A_{k}^{[p]} $\;
  }}  
  $V_{I, \delta(\lambda)}^{[p]}$ = $\mathbb{E}_{A^{[p]}} [\delta (I \odot m) \cdot C_{m}(\lambda)]$\;
}
 SISE explanation: $Y_{I,\delta(\lambda)}$ = $\zeta (V_{I, \delta(\lambda)}^{[p]})$ \\
 \Output{A 2D explanation map $Y_{I,\delta(\lambda)}$.}
 \caption{Ada-SISE: Adaptive Semantic Input Sampling for Explanation}
 \label{alg:Ada-SISE}
\end{algorithm}

\begin{figure}[ht]
     \centering
     \includegraphics[width=0.95\linewidth]{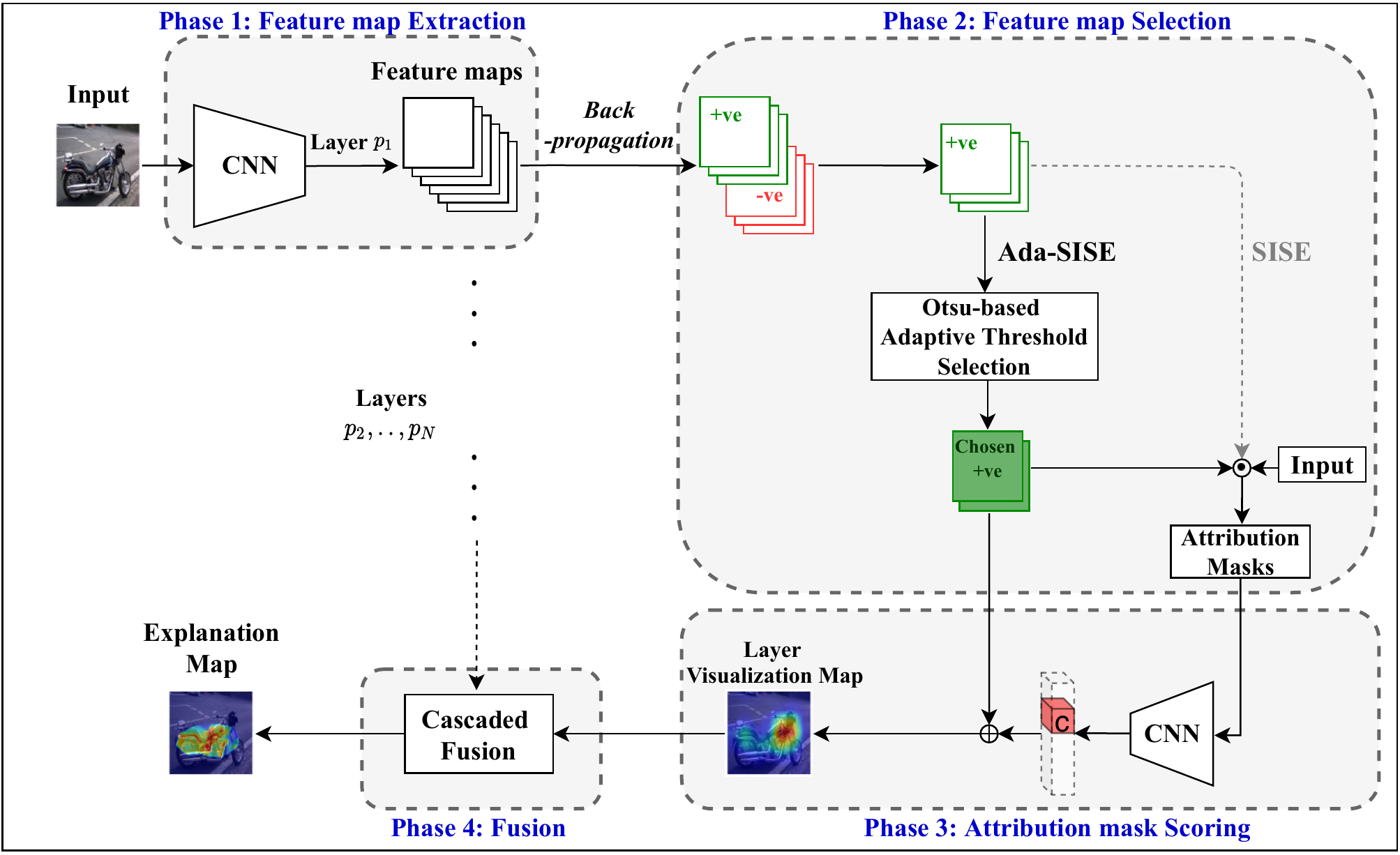}
     \caption{Architecture of the proposed Ada-SISE XAI method. }
     \label{fig:arch}
 \end{figure}
 
To tune the strictness of feature map selection adaptively for each of the layers, we employ an Otsu-based framework \cite{Otsu}.
For a selected layer $p$, we reach the set of feature maps $F_k^{[p]}$ and their corresponding gradient values $\sigma_k^{[p]}$, and determine its maximum as $\rho^{[p]}$. Denoting the normalized gradient values for the feature maps as $\upsilon_k^{[p]}=\frac{\sigma_k^{[p]}}{\rho^{[p]}}$, we define the set of positive-gradient feature maps as:
\begin{equation}
    \Upsilon^{[p]} \equiv {\Upsilon^{[p]}}^{+}= \{\upsilon_k^{[p]} > 0 | k\in\{1,...,M^{[p]}\} \}
\end{equation}
where $M^{[p]}$ is the number of feature maps extracted from layer $p$. Otsu's method is applied to the set of positive-gradient feature maps to implement an updated threshold on them, based on the histogram of their average gradient scores. Assuming $\Upsilon^{[p]}(i) \forall i \in \{1,...,|\Upsilon^{[p]}|\}$ to be the $i$-th
value in $\Upsilon^{[p]}$, we can formulate the mean value of the masks with less/more gradient values than $\Upsilon^{[p]}(i)$ respectively, as follows:
\begin{equation}
    \omega^{[p]}_L(i)=\frac{\sum_{j=1}^{i} (\Upsilon^{[p]}(j))}{i} \times |\Upsilon^{[p]}|
\end{equation}
\begin{equation}
    \omega^{[p]}_H(i)=\frac{\sum_{j=i}^{|\Upsilon^{[p]}|} (\Upsilon^{[p]}(j))}{|\Upsilon^{[p]}|-i}\times |\Upsilon^{[p]}|
\end{equation}
 
If we set $\mu=\Upsilon^{[p]}(i)$ to divide the set of positive-gradient feature maps into two low and high subsets, the inter-class variance of these sets are calculated as follows:
\begin{equation}
    \tau^{[p]}(i)= \omega^{[p]}_L(i) \times \omega^{[p]}_H(i) \times \bigg[ \frac{|\Upsilon^{[p]}|-i}{|\Upsilon^{[p]}|}-\frac{i}{|\Upsilon^{[p]}|} \bigg]^2
\end{equation}
which can be simplified as:
\begin{equation}
    \tau^{[p]}(i)= \omega^{[p]}_L(i) \times \omega^{[p]}_H(i) \times \bigg[ \frac{|\Upsilon^{[p]}|-2i}{|\Upsilon^{[p]}|} \bigg]^2
    \label{eq:obt}
\end{equation}
According to \cite{Otsu}, minimizing the intra-class variance for both classes simultaneously is equivalent to maximizing the inter-class variance in equation \eqref{eq:obt}. Hence, we can identify the most deterministic feature maps in each layer by applying a threshold which maximizes the inter-class variance accordingly:
\begin{equation}
    \mu^{[p]}=\Upsilon^{[p]}\bigg( \argmax_{j\in \{1,...,|\Upsilon^{[p]}|\}} \big( \tau^{[p]}(j) \big) \bigg)
    \label{eq:thres}
\end{equation}

The $\argmax$ operation in equation \eqref{eq:thres} is achieved by a simple search method. If the number of feature maps derived from a layer is not noticeably large, and if some of these feature maps are discarded as negative-gradient activation maps, a simple search method would not add any significant additional complexity to SISE framework. We term our method Ada-SISE and show its architecture in Fig. \ref{fig:arch} and its methodology in Algorithm \ref{alg:Ada-SISE}.

\section{RESULTS}

To compare Ada-SISE's performance abreast with SISE, experiments were performed on the test set of the Pascal VOC 2007 dataset \cite{PASCALVOC}. Two pre-trained models, a shallow VGG16 (with a test accuracy of 87.18\%) and a residual ResNet-50 network (with 87.96\% test accuracy), are directly loaded from the TorchRay library \cite{petsiuk2018rise} to replicate the original experimentation setup. As it was reported in \cite{sattarzadeh2020explaining} that SISE meets or outperforms most of the state-of-the-art XAI methods like Grad-CAM \cite{GradCAM}, RISE \cite{petsiuk2018rise} and Score-CAM \cite{ScoreCAM}, we restrict our comparisons only with Extremal Perturbation \cite{fong2019understanding} (as it is one of the sophisticated perturbation-based methods) and SISE.

\subsection{Benchmark Analysis}

\begin{table}[ht]
 \centering
 \begin{tabular}{c c c c c}
 \toprule 
  & \multirow{2}{*}{\textbf{Metric}} & \textbf{Extremal} & \multirow{2}{*}{\textbf{SISE}}  & \textbf{} \multirow{2}{*}{\textbf{Ada-SISE}}\\
 & & \textbf{Perturbation} & & \\
 \midrule
 \multirow{5}{*}{\STAB{\rotatebox[origin=c]{90}{\textbf{VGG16}}}}
 & \textbf{EBPG}& \textbf{61.19} & 60.54 & 60.79 \\
  & \textbf{Bbox} & 51.2 & 55.68 & \textbf{55.73}\\
 & \textbf{Drop\%} & 43.9 & \textbf{38.40} & 38.87\\
  & \textbf{Increase\%} & 32.65 & 37.96 & \textbf{38.25}\\
  \cmidrule(lr){2-5}
  & \textbf{Run-time} (s) & 87.42 & 5.96 & \textbf{4.23} \\
 \midrule
\multirow{5}{*}{\STAB{\rotatebox[origin=c]{90}{\textbf{ResNet-50}}}}
 & \textbf{EBPG} & 63.24 & 66.08 & \textbf{66.4} \\
  & \textbf{Bbox} & 52.34 & 61.59 & \textbf{61.77} \\
  & \textbf{Drop\%} & 39.38 & \textbf{30.92} & \textbf{30.92} \\
  & \textbf{Increase\%} & 34.27 & 40.22 & \textbf{40.75} \\
  \cmidrule(lr){2-5}
  & \textbf{Run-time} (s) & 78.37 & 9.21 & \textbf{6.29} \\
 \bottomrule
 \end{tabular}
 \caption{Results of benchmark evaluation of Ada-SISE on pre-trained models on the PASCAL VOC 2007 \cite{PASCALVOC} dataset.}
 \label{tab: gt_metrics}
\end{table}

\begin{figure}[ht]
     \centering
     \includegraphics[width=0.8\linewidth]{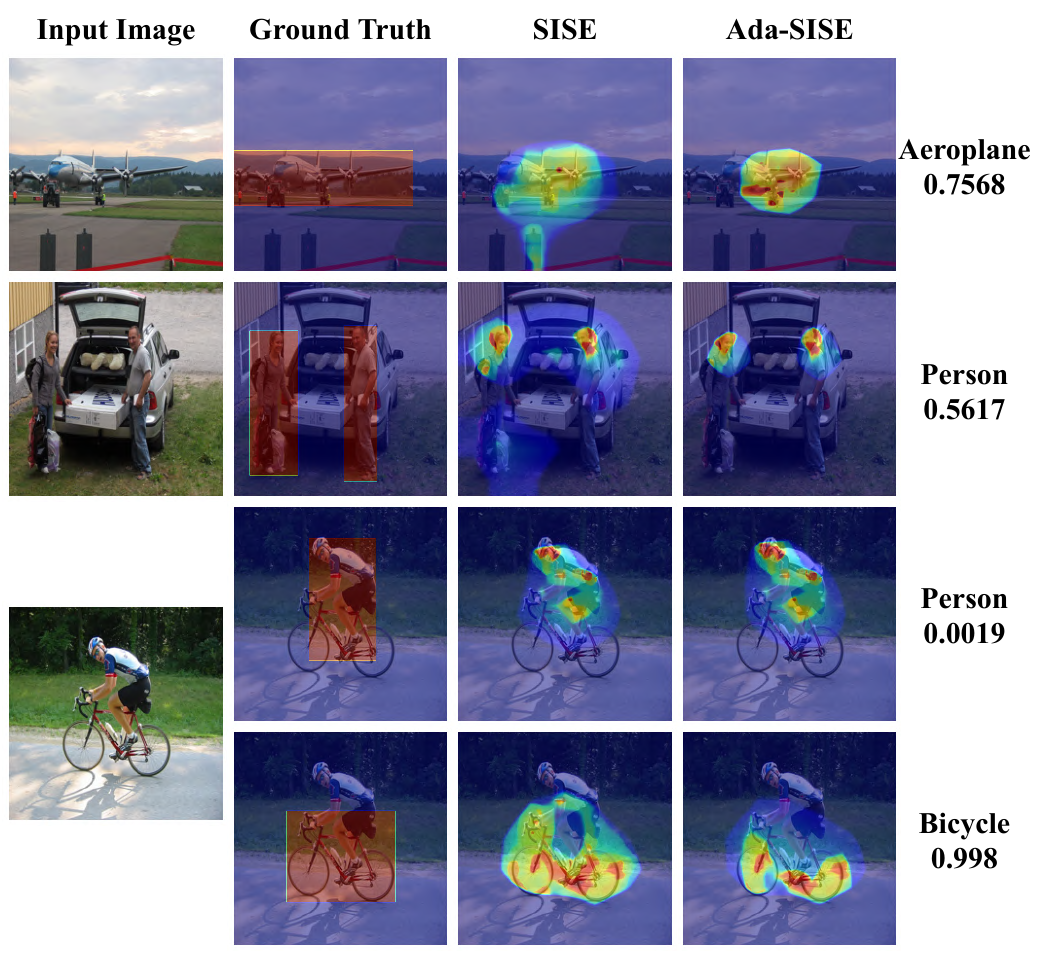}
     \caption{Comparison of Ada-SISE with SISE \cite{sattarzadeh2020explaining} on a ResNet-50 model with images from Pascal VOC 2007 dataset \cite{PASCALVOC} demonstrating their class-discriminative explanation ability.}
     \label{fig:qual}
 \end{figure}

Table \ref{tab: gt_metrics} shows the benchmark evaluation of Ada-SISE concerning various metrics and their execution time. As the depicted results are achieved through the same experimental setup as SISE paper, the readers can refer to \cite{sattarzadeh2020explaining} to infer further head-to-head comparison of Ada-SISE with other state-of-the-art methods. Energy-Based Pointing Game (EBPG) \cite{ScoreCAM} and Bbox \cite{schulz2020restricting} use the ground-truth annotations available to determine the precision of an XAI algorithm. Concurrently, Drop and Increase rates \cite{AblationCAM} measure the contribution of pixels captured in the explanation map towards the model's predictive accuracy. Ada-SISE outperforms SISE in almost all of the metrics\footnote{For each metric in Table \ref{tab: gt_metrics}, the best is shown in bold. Besides Drop\% and run-time (in seconds), the higher is better for all other metrics.} while executing about 30\% faster. Fig. 
\ref{fig:qual} compares the explanation maps qualitatively on a ResNet-50 model and shows the ground-truth class and their annotations along with the model's corresponding confidence score for each image.  

\subsection{Run-time Analysis}

  \begin{figure}[ht]
     \centering
     \includegraphics[width=0.7\linewidth]{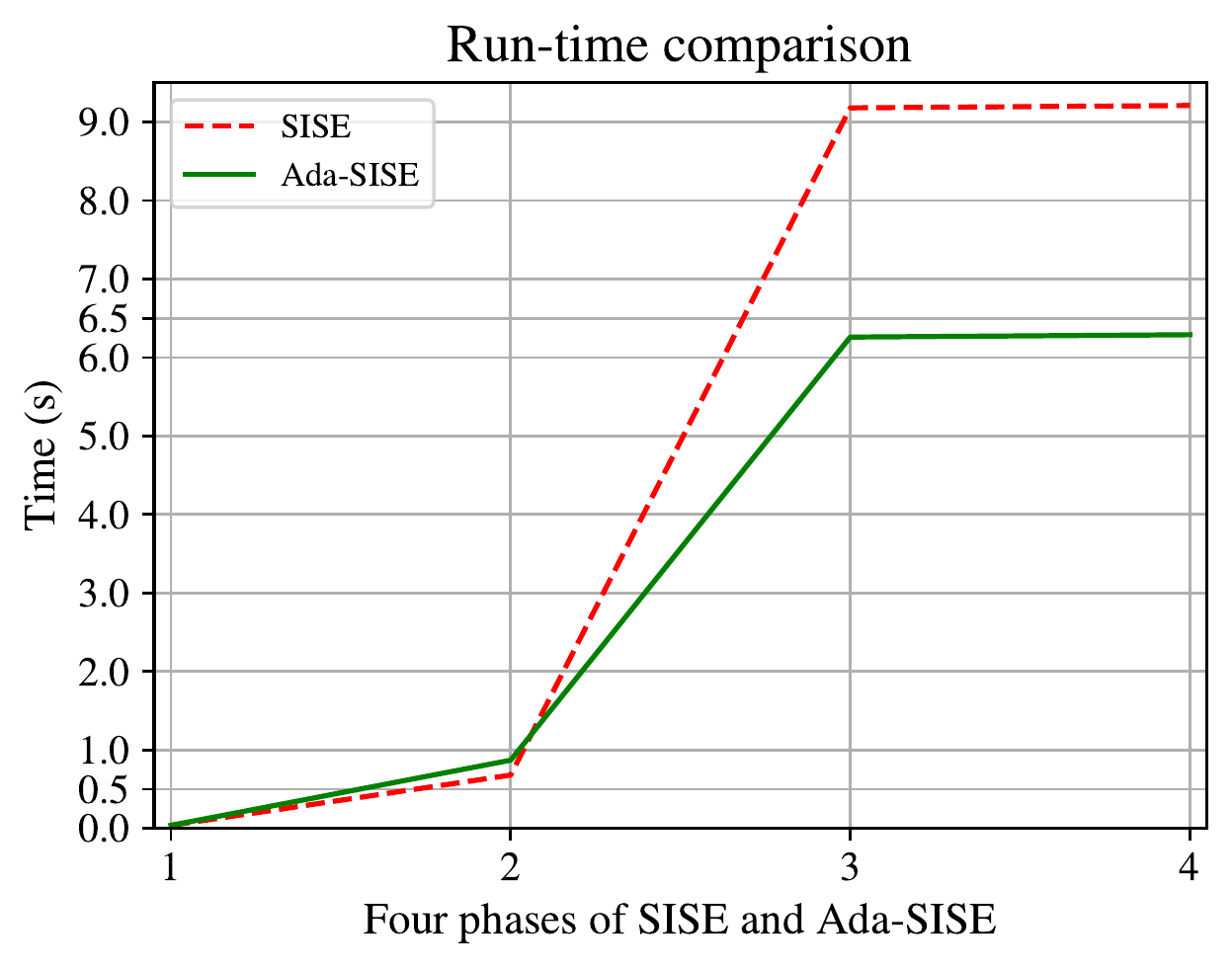}
     \caption{Comparison of the average run-times of Ada-SISE with SISE on a sample of images with a ResNet-50. }
     \label{fig:run-time}
 \end{figure}

The bottleneck in SISE's run-time is its third phase, where too many positive gradient feature maps are feed-forwarded to compute their weights for scoring. As Ada-SISE chooses only a fraction of them that it considers crucial, our algorithm's run-time is reduced significantly in the scoring phase. 

Fig. \ref{fig:run-time} shows the comparison of run-times, where it can be noted that Ada-SISE executes under 6.3 seconds while SISE takes about 9.21 seconds. The small rise in the execution time at the second phase of Ada-SISE is the effect of our proposed adaptive thresholding procedure. The reported numbers are the average of experimentation performed over 100 random images from the Pascal VOC dataset on an NVIDIA Tesla T4 GPU with 16 GB of RAM.

\subsection{Discussion}
\begin{table}[ht]
 \centering
 \begin{tabular}{c c c c c c}
 \textbf{} & \textbf{$p_{1}$} & \textbf{$p_{2}$} & \textbf{$p_{3}$} & \textbf{$p_{4}$} & \textbf{$p_{5}$} \\
 \toprule
 \textbf{No. of feature} & \multirow{2}{*}{64} & \multirow{2}{*}{256} & \multirow{2}{*}{512} & \multirow{2}{*}{1024} & \multirow{2}{*}{2048} \\
 \textbf{maps available} & & & & &\\
 \midrule
 \textbf{SISE} & 31 & 130 & 262 & 515 & 1008 \\
 \midrule
 \textbf{Ada-SISE} & 26 & 114 & 179 & 420 & 551 \\
 \bottomrule
 \end{tabular}
 \caption{The number of feature maps chosen by Ada-SISE (on average) over the five pooling layers of a ResNet-50 compared with that of SISE and the corresponding number of available maps.}
 \label{tab:discussion}
\end{table}


The number of feature maps selected for each pooling layer $p_{l}$ of the network was recorded over a data sample of 500 images from the Pascal dataset, averaged, and reported in Table \ref{tab:discussion}. As the deeper layers contribute more feature maps, it can be noticed that Ada-SISE chooses only a fraction of them, justifying the run-time reduction after the second phase. This validates our claim that by neglecting comparatively lower gradient values, dominant feature maps that contribute more towards a prediction can be extracted without compromising the explanation quality. Although an ablation study could be performed to identify a suitable value for $\mu$ by fine-tuning SISE through extensive experiments, this solution would be profoundly dependent on the training data and would be brittle when expanded to new unseen data. Therefore, Ada-SISE generalizes SISE to be scaled for any application.

\section{CONCLUSION}

In this work, we propose Ada-SISE as an improvement to the recent SISE method that makes it a fully automated XAI algorithm. We also report a reduction in run-time and an overall improvement in the benchmark analysis without losing its visual explanation quality. Such identified important features would be adopted in future works to analyze a model's behavior by studying its effect on the model's prediction when replaced with noises or other classes' attributions.


\bibliographystyle{IEEEbib}
\bibliography{refs}

\begin{thebibliography}{10}

\bibitem{fan2020interpretability}
Fenglei Fan, Jinjun Xiong, and Ge~Wang,
\newblock ``On interpretability of artificial neural networks,''
\newblock {\em arXiv preprint arXiv:2001.02522}, 2020.

\bibitem{arrieta2020explainable}
Alejandro~Barredo Arrieta, Natalia D{\'\i}az-Rodr{\'\i}guez, Javier Del~Ser,
  Adrien Bennetot, Siham Tabik, Alberto Barbado, Salvador Garc{\'\i}a, Sergio
  Gil-L{\'o}pez, Daniel Molina, Richard Benjamins, et~al.,
\newblock ``Explainable artificial intelligence (xai): Concepts, taxonomies,
  opportunities and challenges toward responsible ai,''
\newblock {\em Information Fusion}, vol. 58, pp. 82--115, 2020.

\bibitem{adebayo2016fairml}
Julius~A Adebayo et~al.,
\newblock {\em FairML: ToolBox for diagnosing bias in predictive modeling},
\newblock Ph.D. thesis, Massachusetts Institute of Technology, 2016.

\bibitem{bellamy2019ai}
Rachel~KE Bellamy, Kuntal Dey, Michael Hind, Samuel~C Hoffman, Stephanie Houde,
  Kalapriya Kannan, Pranay Lohia, Jacquelyn Martino, Sameep Mehta,
  A~Mojsilovi{\'c}, et~al.,
\newblock ``Ai fairness 360: An extensible toolkit for detecting and mitigating
  algorithmic bias,''
\newblock {\em IBM Journal of Research and Development}, vol. 63, no. 4/5, pp.
  4--1, 2019.

\bibitem{sattarzadeh2020explaining}
Sam Sattarzadeh, Mahesh Sudhakar, Anthony Lem, Shervin Mehryar, KN~Plataniotis,
  Jongseong Jang, Hyunwoo Kim, Yeonjeong Jeong, Sangmin Lee, and Kyunghoon Bae,
\newblock ``Explaining convolutional neural networks through attribution-based
  input sampling and block-wise feature aggregation,''
\newblock {\em arXiv preprint arXiv:2010.00672}, 2020.

\bibitem{IntegGrad}
Mukund Sundararajan, Ankur Taly, and Qiqi Yan,
\newblock ``Axiomatic attribution for deep networks,''
\newblock {\em arXiv preprint arXiv:1703.01365}, 2017.

\bibitem{simonyan2013deep}
Karen Simonyan, Andrea Vedaldi, and Andrew Zisserman,
\newblock ``Deep inside convolutional networks: Visualising image
  classification models and saliency maps,''
\newblock {\em arXiv preprint arXiv:1312.6034}, 2013.

\bibitem{srinivas2019full}
Suraj Srinivas and Fran{\c{c}}ois Fleuret,
\newblock ``Full-gradient representation for neural network visualization,''
\newblock in {\em Advances in Neural Information Processing Systems}, 2019, pp.
  4126--4135.

\bibitem{bach2015pixel}
Sebastian Bach, Alexander Binder, Gr{\'e}goire Montavon, Frederick Klauschen,
  Klaus-Robert M{\"u}ller, and Wojciech Samek,
\newblock ``On pixel-wise explanations for non-linear classifier decisions by
  layer-wise relevance propagation,''
\newblock {\em PloS one}, vol. 10, no. 7, pp. e0130140, 2015.

\bibitem{petsiuk2018rise}
Vitali Petsiuk, Abir Das, and Kate Saenko,
\newblock ``Rise: Randomized input sampling for explanation of black-box
  models,''
\newblock {\em arXiv preprint arXiv:1806.07421}, 2018.

\bibitem{fong2019understanding}
Ruth Fong, Mandela Patrick, and Andrea Vedaldi,
\newblock ``Understanding deep networks via extremal perturbations and smooth
  masks,''
\newblock in {\em Proceedings of the IEEE International Conference on Computer
  Vision}, 2019, pp. 2950--2958.

\bibitem{CAM}
Bolei Zhou, Aditya Khosla, Agata Lapedriza, Aude Oliva, and Antonio Torralba,
\newblock ``Learning deep features for discriminative localization,''
\newblock in {\em Proceedings of the IEEE conference on computer vision and
  pattern recognition}, 2016, pp. 2921--2929.

\bibitem{zhou2014object}
Bolei Zhou, Aditya Khosla, Agata Lapedriza, Aude Oliva, and Antonio Torralba,
\newblock ``Object detectors emerge in deep scene cnns,''
\newblock {\em arXiv preprint arXiv:1412.6856}, 2014.

\bibitem{GradCAM}
Ramprasaath~R Selvaraju, Michael Cogswell, Abhishek Das, Ramakrishna Vedantam,
  Devi Parikh, and Dhruv Batra,
\newblock ``Grad-cam: Visual explanations from deep networks via gradient-based
  localization,''
\newblock in {\em Proceedings of the IEEE international conference on computer
  vision}, 2017, pp. 618--626.

\bibitem{GradCAMPP}
Aditya Chattopadhay, Anirban Sarkar, Prantik Howlader, and Vineeth~N
  Balasubramanian,
\newblock ``Grad-cam++: Generalized gradient-based visual explanations for deep
  convolutional networks,''
\newblock in {\em 2018 IEEE Winter Conference on Applications of Computer
  Vision (WACV)}. IEEE, 2018, pp. 839--847.

\bibitem{ScoreCAM}
Haofan Wang, Zifan Wang, Mengnan Du, Fan Yang, Zijian Zhang, Sirui Ding, Piotr
  Mardziel, and Xia Hu,
\newblock ``Score-cam: Score-weighted visual explanations for convolutional
  neural networks,'' 2020.

\bibitem{Otsu}
N.~{Otsu},
\newblock ``A threshold selection method from gray-level histograms,''
\newblock {\em IEEE Transactions on Systems, Man, and Cybernetics}, vol. 9, no.
  1, pp. 62--66, 1979.

\bibitem{PASCALVOC}
M.~Everingham, L.~Van~Gool, C.~K.~I. Williams, J.~Winn, and A.~Zisserman,
\newblock ``The {PASCAL} {V}isual {O}bject {C}lasses {C}hallenge 2007
  {(VOC2007)} {R}esults,'' 2007.

\bibitem{schulz2020restricting}
Karl Schulz, Leon Sixt, Federico Tombari, and Tim Landgraf,
\newblock ``Restricting the flow: Information bottlenecks for attribution,''
\newblock {\em arXiv preprint arXiv:2001.00396}, 2020.

\bibitem{AblationCAM}
Harish~Guruprasad Ramaswamy et~al.,
\newblock ``Ablation-cam: Visual explanations for deep convolutional network
  via gradient-free localization,''
\newblock in {\em The IEEE Winter Conference on Applications of Computer
  Vision}, 2020, pp. 983--991.

\end{thebibliography}

\end{document}